\pdfoutput=1

\documentclass[11pt]{article}

\usepackage[]{acl}

\usepackage{times}
\usepackage{latexsym}

\usepackage{multirow}
\usepackage{float}
\usepackage{graphicx}
\usepackage{enumitem}

\usepackage[T1]{fontenc}

\usepackage[utf8]{inputenc}

\usepackage{microtype}

\title{Zero-shot Aspect-level Sentiment Classification via Explicit Utilization of Aspect-to-Document Sentiment Composition}

\author{Pengfei Deng, Jianhua Yuan, Yanyan Zhao, Bing Qin \\
    Research Center for Social Computing and Information Retrieval,\\
    Faculty of Computing,\\
    Harbin Institute of Technology, Harbin 150001, China\\
    \{pfdeng, jhyuan, yyzhao, qinb\}@ir.hit.edu.cn
}

\begin{document}
\maketitle
\begin{abstract}
As aspect-level sentiment labels are expensive and labor-intensive to acquire, zero-shot aspect-level sentiment classification is proposed to learn classifiers applicable to new domains without using any annotated aspect-level data. In contrast, document-level sentiment data with ratings are more easily accessible. In this work, we achieve zero-shot aspect-level sentiment classification by only using document-level reviews. Our key intuition is that the sentiment representation of a document is composed of the sentiment representations of all the aspects of that document. Based on this, we propose the AF-DSC method to explicitly model such sentiment composition in reviews. AF-DSC first learns sentiment representations for all potential aspects and then aggregates aspect-level sentiments into a document-level one to perform document-level sentiment classification. In this way, we obtain the aspect-level sentiment classifier as the by-product of the document-level sentiment classifier.  Experimental results on aspect-level sentiment classification benchmarks demonstrate the effectiveness of explicit utilization of sentiment composition in document-level sentiment classification. Our model with only 30k training data outperforms previous work utilizing millions of data.
\end{abstract}

\section{Introduction}

Aspect-based sentiment classification (ASC) aims to predict the sentiment polarity of aspects in a review. For example, in the review ``The food is top notch but the service is heedless'', there are two aspects ``food'' and ``service'', and their sentiment polarities are positive and negative respectively. However, due to the high cost of annotation, annotated datasets are not always available in all domains, which brings obstacles to the application of the model to different domains, resulting in the emergence of zero-shot ASC \cite{Seoh:2021}.

Existing work in zero-shot ASC \cite{Seoh:2021} uses review text to perform post-train on the pre-trained language model and utilize the language model ability of the model to accomplish zero-shot ASC. In some previous works~\cite{He:2018,Zhou:2020,Ke:2020}, it was generally believed that there is a correlation between DSC and ASC, and the knowledge of DSC will facilitate ASC because both of them are aimed at judging the sentiment related to the review. The former wants to predict the sentiment of the whole review (rating in other words), while the latter needs to predict the sentiment of the aspect existing in the review text. Therefore, in these works, they simply add the DSC task as an auxiliary training task into the training process of the model on the ASC task, but they did not point out the specific correlation between ASC and DSC, which also hinders the use of document-level knowledge to complete the zero-shot ASC.

\begin{figure*}
	\centering
	\includegraphics[scale=0.45]{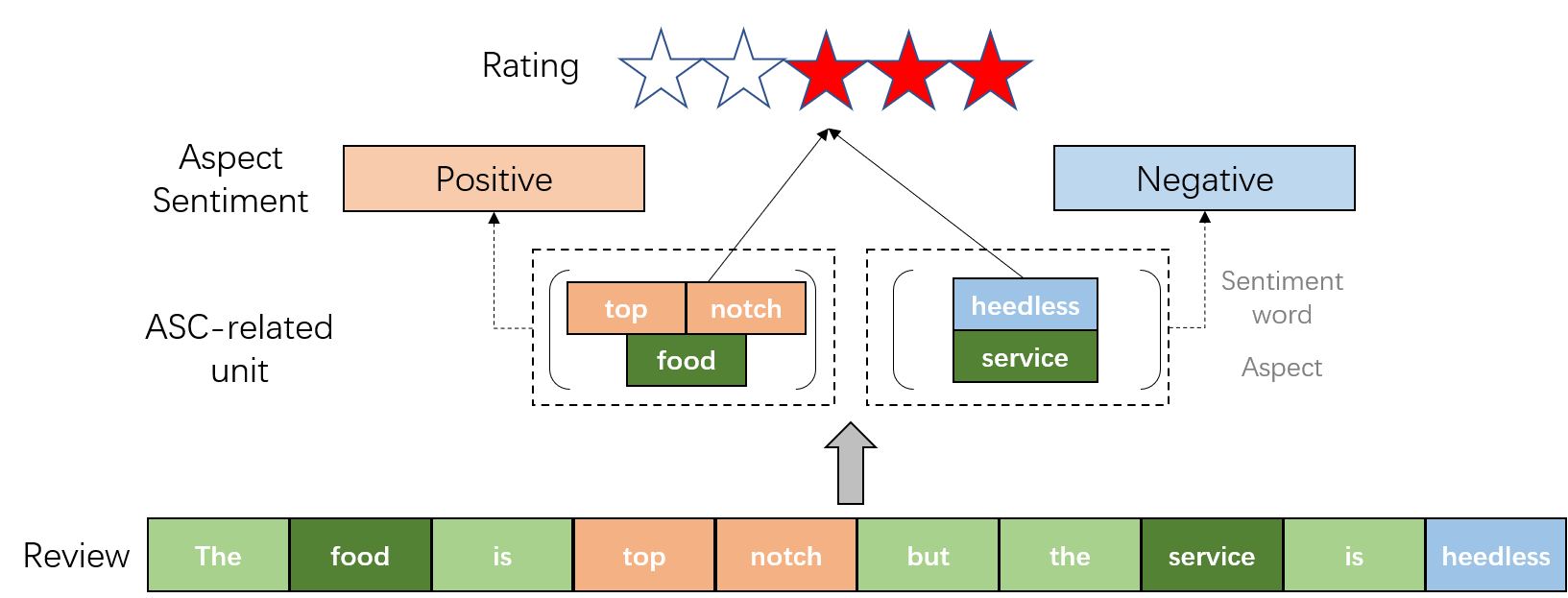}
	\caption{\label{fig:1}The relationship between ASC-related units and the rating label. There are two ASC-related units in the review, (`good', `food') and (`nice', `service'), which together constitute the 5 stars rating label.}
\end{figure*}

We suppose the rating label in a document-level review text could be viewed as a sentiment bond to associate ASC with DSC. The value of a rating label is always composed of fine-grained sentimental clues from multiple ASC-related units contained in the document text. Each ASC-related unit takes an aspect as the core and the sentiment polarity word modifying the aspect as the auxiliary. As shown in Figure~\ref{fig:1}, there are two ASC-related units in ``The food is top notch but the service is heedless'', (`top notch', `food') and (`heedless', `service') respectively, so the aspect sentiment of `food' and `service' is positive and negative respectively, the two ASC-related units compose the 3 stars rating label. According to this, during the process of predicting the rating label of DSC, representations of ASC-related units learned from a large amount of easy-to-obtain document-level data would provide rich sentiment clues for zero-shot ASC.

In this paper, to perform zero-shot ASC with fine-grained sentimental clues obtained from DSC, we propose the method \underline{A}spect \underline{F}ocus – \underline{D}ocument \underline{S}entient \underline{C}lassification (AF-DSC) to form representations of ASC-related units when predicting the rating labels of DSC. Specifically, we train the DSC according to the aspect-document two-step process of forming ASC-related unit representation first and then document-level sentiment representation. In other words, we regard the process of forming ASC-related unit representation as an intermediate step in forming the document-level sentiment label. When using the model for the zero-shot ASC task, we take out the intermediate step individually to obtain the sentiment polarity of the target aspect. In addition, two auxiliary tasks about aspect and sentiment words, named Word Sentiment Prediction and Mask Word Prediction are introduced to perform multi-task learning for boosting the performance of our model in the zero-shot ASC task.

The main contributions of this paper can be summarized as follows:
\textbf{(1)} We explicitly model the relationship between the ASC task and the DSC task and obtain ASC sentiment as the training by-product when training the DSC task.
\textbf{(2)} We design AF-DSC to attain representations of ASC-related units for zero-shot ASC. And two auxiliary tasks are introduced for further improvement in the zero-shot ASC task.
\textbf{(3)} Experimental results show that our model not only outperforms the baseline models but also achieves good performance under the cross-domain setting.

\section{Related Work}

\subsection{Zero-shot Sentiment Analysis}
\label{sect:pdf}
Traditional sentiment analysis tasks are document-level oriented~\cite{Hu:2004,Pang:2008,Liu:2012}. But aspect-based sentiment analysis is aspect-oriented and a more fine-grained task for sentiment analysis~\cite{Pontiki:2014}. Since the dataset size of aspect-based sentiment analysis is generally small, Sun et al.~\shortcite{Sun:2019} and Xu et al.~\shortcite{Xu:2019} use task-related knowledge contained in SQuAD question-answering datasets to handle this problem. There still exists a problem that if there are no annotated datasets available, the existing ASC models are not trainable, which leads to the emergence of zero-shot sentiment analysis.

Many NLP tasks can be formally transformed into the form of natural language inference. Some zero-shot NLP tasks try to use the strong natural language inference ability of the pre-trained language model. Yin et al.,\shortcite{Yin:2019} reformulate text classification tasks as a textual entailment formulation.  In their subsequent work, Yin et al., \shortcite{Yin:2020} apply natural language inference tasks to solve two few-shot NLP tasks (Question Answering and Coreference Resolution). In the work of Sainz et al., \shortcite{Sainz:2021}, they use natural language inference to complete zero-shot and few-shot relation extraction. In the work of Seoh et al., \shortcite{Seoh:2021}, our main baseline models, natural language inference model, and language model are also used to complete the aspect sentiment classification task under zero-shot setting and few-shot setting. Shu et al. \shortcite{Shu:2022} cast three ABSA tasks (Aspect Extraction, 
Aspect Sentiment Classification, End-to-End Aspect-Based Sentiment Analysis) into natural language inference (NLI) to allow the zero-shot transfer.

However, because there is little sentiment-related content in the current natural language inference datasets, it is difficult for the natural language inference model to better complete ASC under the zero-shot setting. In contrast, DSC is closer to the ASC in sentiment. Therefore, in our work, we use document-level sentiment data to complete the zero-shot ASC.

\subsection{Transfer Document Knowledge To ASC}

To solve aspect-based sentiment classification, a series of neural network models have been proposed. However, due to the high cost of data acquisition, the existing aspect-based sentiment classification datasets are small. To solve this problem, He et al. \shortcite{He:2018} try to integrate document-level sentiment knowledge into the training of the aspect-based sentiment classification model for the first time. In their subsequent work, He et al. \shortcite{He:2019} also integrate document-level related information into the model as training tasks, and both of them receive performance gains. Chen and Qian \shortcite{Chen:2019} propose a novel Transfer Capsule Network model to transfer sentence-level sentiment knowledge from DSC to ASC. In their work, they try to incorporate two kinds of sentiment preference information (intra-aspect consistency and inter-aspect tendency) in a document for remedying the information deficiency problem in ASC. In the work of Zhou et al. \shortcite{Zhou:2020} and Ke et al. \shortcite{Ke:2020}, they integrate the DSC task as a training task into the post-train of the pre-trained language model on the target domain corpus. In the work of Li et al., \shortcite{Li:2019}, the aspect category sentiment classification data is used to enhance the ability of aspect-based sentiment classification through the coarse to fine task transfer.

Inspired by the above works, our work tries to integrate DSC into the zero-shot ASC. But the above works do not point out the explicit relation between the ASC task and the DSC task. In our work, we explicitly model the relationship between them and propose that the document-level sentiment is synthesized by the ASC-related units in the review.

\section{Methods}

\begin{figure*}
	\centering
	\includegraphics[scale=0.45]{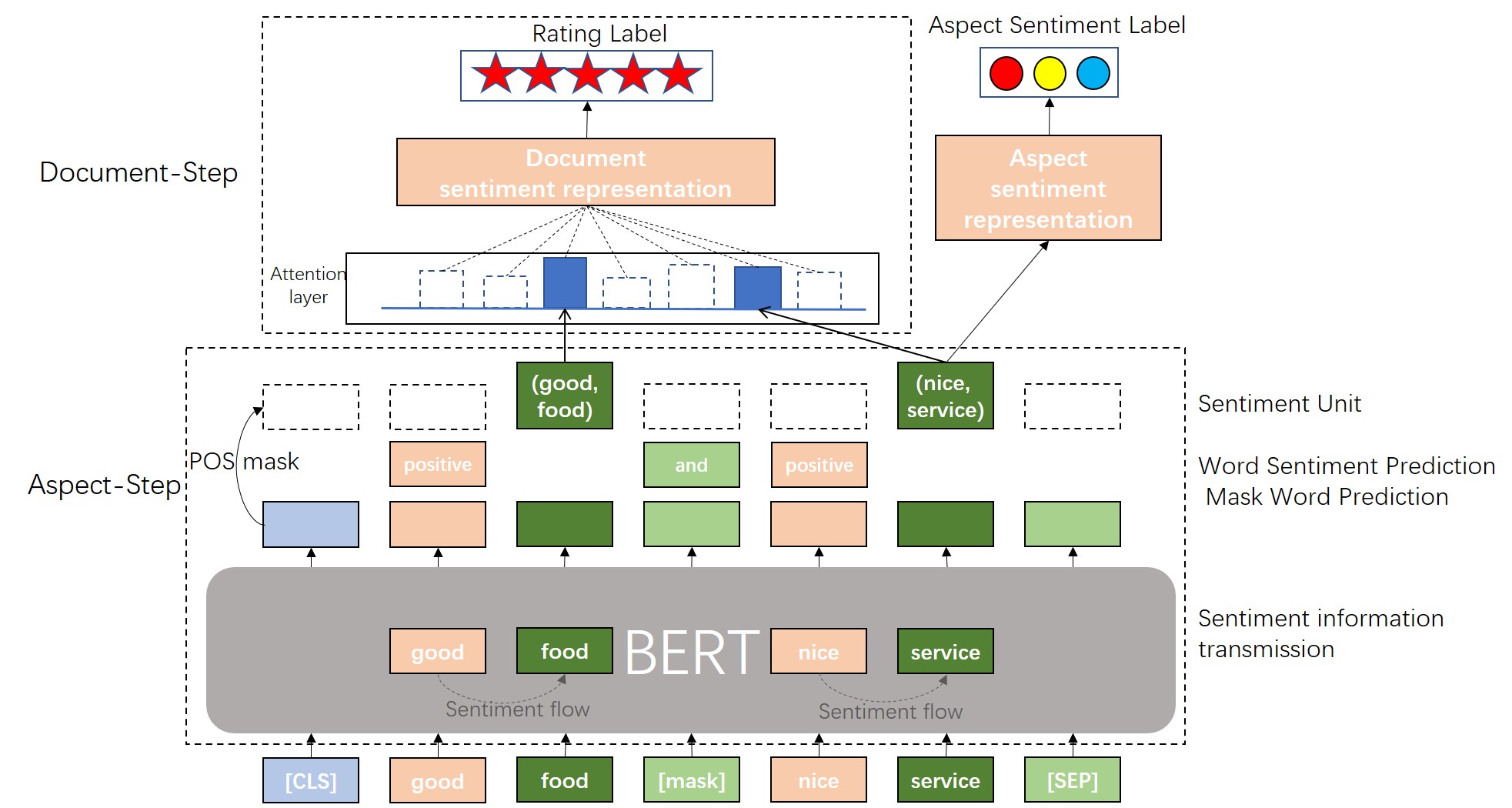}
	\caption{\label{fig:2}Overview of our model. The model consists of Aspect-Step which forms the ASC-related unit representation and the Document-Step which forms the document sentiment representation. We further use Word Sentiment Prediction and Mask Word Prediction tasks to get a better ASC-related unit representation.}
\end{figure*}

As shown in Figure~\ref{fig:2}, the model constructs an aspect-document two-step process that gets aspect sentiment first and then forms document-level sentiment. We get the latent aspects and ASC-related unit representations by rule-based extraction and finally form document-level sentiment representation.

\subsection{Rule-based Aspect Acquisition}

Because there is no annotation about the aspects in the document-level sentiment classification datasets, we roughly extract the potential aspects by seeking rules. Intuitively, most aspects are composed of nouns. Therefore, we roughly consider all words whose part of speech label is noun are the potential aspects.

For each document level sentiment review $ s $, we use the Spacy library \cite{Honnibal:2017} to segment words and get the part of speech tag of the word. Since the word tokenizer used in the BERT model is different from that used in the Spacy, we use pytokenizations\footnote{\url{https://github.com/explosion/tokenizations}} to align the results after word tokenization. So far, we have obtained the part of speech tags $ \left[p_1, p_2, \ldots, p_n\right] $ corresponding to the BERT word tokenization result, $ n $ is the number of tokens. We then obtain the mask of potential aspects $ \left[m_1, m_2, \ldots, m_n\right] $, where $ m_i = 1 $ if $ p_i \in D_{noun} $ indicates it is a potential aspect, otherwise $ m_i = 0 $ , $ D _{noun} $ is a collection of noun part of speech tags. Since we use Spacy to acquire part of speech tag, we have $ D_{noun} = \left\{NOUN,PROPN\right\} $ means noun and proper noun respectively.

\subsection{Composing Aspect Sentiments To Document Sentiment}

Using rule-based extraction, we roughly obtain the aspect in the ASC-related unit. To obtain the information on sentiment polarity words which is another element in the ASC-related unit, we use the relationship between the aspect and sentient words established by the self-attention mechanism in the transformer layer and then transfer the sentiment-related information in sentiment words to the hidden state representation of aspect word to form the ASC-related unit representation. Firstly, we take the output hidden state of the BERT model corresponding to the extracted potential aspect as the ASC-related unit representation. Then, we use an attention layer on the ASC-related unit representations, and aggregate them to form the document-level sentiment representation $ h_{rating} $. The calculation process is given by the following formulas:

\begin{equation}
	{\widehat \beta _i} = {t^T}{h_i} + b
\end{equation}
\begin{equation}
	{\beta _i} = {\hat \beta _i} * {m_i} + (1 - {m_i}) * ( - inf )
	\label{con:pos_mask}
\end{equation}
\begin{equation}
	{\alpha _i} = \frac{{\exp ({\beta _i})}}{{\sum\nolimits_{j = 1}^n {\exp ({\beta _j})} }}
\end{equation}
\begin{equation}
	{h_{rating}} = \sum\limits_{i = 1}^n {{\alpha _i}{h_i}}
\end{equation}
where $ h_i $ is the hidden state output vector of BERT corresponding to the ith word, $ n $ is the number of words, $t$ and $b$ are the trainable parameters in the attention layer, $-inf$ means we do not pay attention to those words which are not potential aspects. 

In the beginning, there is only aspect-relevant information in the ASC-related unit representation, but the model needs to acquire sentiment information to predict the document level sentiment, the sentiment information about the aspect will be transferred from the sentiment polarity word to the ASC-related unit representation according to the relationship between the words established by the self-attention mechanism in the pre-trained language model. In this way, the ASC-related unit representation contains the sentiment information related to the target aspect.

\subsection{Enhancing Aspect Sentiment Representation Tasks}
In addition to the document-level sentiment classification task, we also introduce Mask Word Prediction and Word Sentiment Prediction tasks to conduct multi-task learning for enhancing the performance of the model. Our goals are: (1) Word Sentiment Prediction task makes the model aware of the sentiment polarity of sentiment words (2) Mask Word Prediction task further enhances the relationship between the aspect and sentiment word (3) Document Level Sentiment Classification task makes the sentiment information in sentiment word transmit to the ASC-related unit representation.

\begin{itemize}[leftmargin=*]
	\item \textbf{Word Sentiment Prediction}: For the Word Sentiment Prediction task, we use the opinion lexicon \cite{Hu:2004} to get the sentiment of words. We label the words with positive and negative sentiment as ``P'' and ``N'' respectively, and let the model predict the sentiment polarity of these sentiment words during the training process. For the rest of the words, we do not predict their sentiment, because we need to transfer the sentimental information to the hidden state of the aspects, which hardly appear in the opinion lexicon, and the label is likely to be neutral. If we predict it, it may hinder the process of transferring the sentiment information to the hidden state of the aspects. The loss function of our Word Sentiment Prediction task is as follows, where $ {polar}_i $ is the sentiment polarity of the ith word, $ S_i $ is the sign of whether the ith word is a sentiment word. If it is a sentiment word, $ S_i=1 $, otherwise is 0:
	
	\begin{equation}
	    {L_{wsp}} =  - \sum\limits_{i = 1}^n {{S_i} \cdot \log \left( {P\left( {pola{r_i}|{h_i}} \right)} \right)}
    \end{equation}
    
	\item \textbf{Mask Word Prediction}: We hope the model can better establish the relationship between the aspect and sentiment polarity words, we increase the mask probability of nouns, proper nouns, adjectives, and adverbs from 15$ \% $ in the BERT model to 30$ \% $, because nouns and proper nouns are more likely to be potential aspects, while adjectives and adverbs are more likely to be potential sentiment polarity words. The mask probability of other words remains unchanged, which is still 15$ \% $. The loss function of our Mask Word Prediction task is as follows, where $ x_i $ is the ith word:
 
    \begin{equation}
 	    {L_{mwp}} =  - \sum\limits_{i = 1}^n {\log \left( {P\left( {{x_i}|{h_i}} \right)} \right)} 
    \end{equation}
\end{itemize}

\subsection{Sentiment Classification}

After obtaining the document level sentiment representation $ h_{rating} $, we put it into the document level sentiment classifier to get the rating label $ l $. The loss function of the Rating Prediction task is as follows:

\begin{equation}
    {L_{rating}} =  - \log \left( {P\left( {l|{h_{rating}}} \right)} \right) 
\end{equation}

Our joint training loss function is given as follows. To reduce the influence of the mask word prediction task, we Set $ \lambda $ to 0.01.

\begin{equation}
    L = {L_{wsp}} + \lambda  \cdot {L_{mwp}} + {L_{rating}}
\end{equation}

To predict the sentiment of the aspect, the output hidden states corresponding to the tokens of the aspect are averaged, and then sent to the document level sentiment classifier, and the result is mapped to the aspect sentiment label, that is, predicate result $< 3$, $> 3$, and $= 3$ are considered as negative, positive, and neutral respectively.

\section{Experiments}
\subsection{Datasets}

\begin{itemize}[leftmargin=*]
	\item \textbf{Document Sentiment Classification Dataset}: The document-level sentiment classification datasets used in our work from He \shortcite{He:2018}. It contains two datasets, electronics large and yelp large. Each dataset contains 30K instances with the rating tag.
	
	\item \textbf{Aspect-based Sentiment Classification Dataset}: We test the performance of our model on SemEval2014 task 4 \cite{Pontiki:2014}, the most widely used aspect-based sentiment classification datasets, which contains the data from the laptop domain (Lap14) and restaurant domain (Res14). Similar to the previous work, we remove the samples with conflict labels. For the samples with multiple aspects in the review text, we divide them into different samples.
	
	\item \textbf{Aspect Robustness Test Set}: To test whether our AF-DSC model only regards document sentiment as aspect sentiment. We test on the Aspect Robustness Test Set (ARTS) proposed by Xing et al., \shortcite{Xing:2020}. Based on the original SemEval2014 datasets, the author obtains new datasets (ARTS-Res and ARTS-Lap) for testing the robustness of the aspect-based sentiment analysis models by reversing the sentiment polarity of the target aspect (REVTGT), reversing the sentiment polarity of the non-target aspect (REVNON), and adding more aspects opposite to the target aspect (ADDDIFF). In the constructed dataset, the sentiment polarity of the non-target aspects is usually opposite to the target aspect, so the document-level sentiment will be different from the target aspect sentiment. If only using the document level sentiment as the aspect sentiment, a wrong prediction will be made.
\end{itemize}

\subsection{Implementation details}
During training, we use Adam optimizer. The batch size is set to 32, the learning rate is set to 2e-5, the warm ratio is set to 0.1, and the maximum gradient of the model is set to 1.0. We train our models for 2 epochs on the yelp large dataset and 3 epochs on the electronics large dataset. All the results we reported are averaged by 5 runs with different random seeds.

\subsection{Baselines}
We mainly compare our AF-DSC model with the work about zero-shot ASC of Seoh et al. \shortcite{Seoh:2021}. To prove that our work is indeed different from directly treating document-level sentiment as aspect sentiment, we compare the AF-DSC model with the BERT-sent model.

\begin{itemize}[leftmargin=*]
	\item BERT LM\cite{Seoh:2021}: First, perform post-train on the large-scale review corpus, and then construct a prompt like ``I felt the {aspect} was [mask]'', let the model predict the word at [mask], the probability of predicting positive, neutral, and negative sentiment is proportional to the probability of predicting good, ok, and bad, respectively.
	\item GPT-2 LM\cite{Seoh:2021}: Similar to BERT LM, except that the model is changed from the BERT model to GPT-2 model. BERT uses mask word prediction when pretraining, while GPT-2 uses causal language model task instead.
	\item BERT NLI\cite{Seoh:2021}: This work transforms the ASC into a natural language inference task, takes review as the premise, and constructs hypotheses similar to ``the {aspect} is good'' and ``the {aspect} is bad.'' The sentiment of the aspect is decided by the entailment probability between the premise and the hypothesis through the natural language inference model. This work is based on the BERT model trained on the natural language inference dataset.
	\item BERT-sent: We use the output hidden state corresponding to [CLS] as the document sentiment representation. For the aspect-based sentiment classification task, we just feed the review into BERT to get the document-level sentiment label and directly transform it into the aspect sentiment. We also perform the Word Sentiment Prediction task and Mask Word Prediction task during training.
\end{itemize}

\subsection{Main Results}

\begin{table*}
\setlength{\tabcolsep}{6mm}{
\begin{center}
\begin{tabular}{ccccc}
		\hline
		\multirow{2}{*}{Model} & \multicolumn{2}{c}{Res14} & \multicolumn{2}{c}{Lap14} \\ \cline{2-5} 
		& Acc.         & F1         & Acc.         & F1         \\ \hline
		BERT   LM              & 70.86       & 48.17       & 63.58       & 46.17       \\ \hline
		GPT-2 LM               & \textbf{71.40}       & 45.53       & 60.45       & 39.59       \\ \hline
		BERT   NLI             & 61.79       & 57.93       & 58.93       & 54.91       \\ \hline
		ours                   &             &             &             &             \\ \hline
		BERT-sent              & 70.22       & 60.60       & 64.30       & 55.88       \\ \hline
		AF-DSC                 & 69.61       & \textbf{61.69}       & \textbf{65.41}       & \textbf{57.81}       \\ \hline
	\end{tabular}
	\end{center}}
	\caption{\label{table:Main-Results-table} Results of our model and baselines. Acc. and F1 refer to accuracy and macro F1, respectively. The results of BERT LM, GPT-2 LM, and BERT NLI are obtained from the source paper. }
\end{table*}

As shown in Table~\ref{table:Main-Results-table}. In the baseline models, the works based on the language model (BERT LM, GPT-2 LM) perform well in accuracy but get bad results in the macro-F1 score. This may caused by the choice of words predicted by the language model. The word ``OK'' tends to be positive, but use to represent neutral polarity, resulting in the error in the neutral sentiment prediction, and the low macro-F1 score. The work based on natural language inference (BERT NLI) gets poor results in accuracy but achieves better results in the macro-F1 score. This could be the result about the lack of sentiment-related content in the natural language inference dataset .

As we can see, our model exceeds the previous work of Seoh et al., \shortcite{Seoh:2021}. This shows that the rating tag has a rich sentimental connotation and can effectively enhance the sentiment discrimination ability of the model. Therefore, we only used 30k document-level sentiment classification data and achieve better results than using millions of review data for post-train. BERT-sent can achieve good results, which shows that the sentiment polarities of different aspects in the review are the same in most cases. Our AF-DSC model exceeds BERT-sent in Macro-F1 score, which shows that our model can distinguish aspect sentiment from document sentiment data.

\subsection{Cross Domain Analysis}

\begin{table}[ht]
	\begin{center}
	\footnotesize 
	\begin{tabular}{cccccc}
		\hline
		\multirow{2}{*}{Model} & \multirow{2}{*}{In/Cross} & \multicolumn{2}{c}{Res14} & \multicolumn{2}{c}{Lap14} \\ \cline{3-6} 
		&       & Acc.   & F1   & Acc.   & F1   \\ \hline
		\multirow{2}{*}{BERT-sent} & In    & 70.22 & 60.60 & 64.30 & 55.88 \\ \cline{2-6} 
		& Cross & 70.26 & 55.08 & 66.99 & 63.38 \\ \hline
		\multirow{2}{*}{AF-DSC}    & In    & 69.61 & 61.69 & 65.41 & 57.81 \\ \cline{2-6} 
		& Cross & 70.24 & 56.16 & 67.53 & 64.76 \\ \hline
	\end{tabular}
	\end{center}
	\caption{\label{table:Cross-Domain-Analysis-table} Cross-domain performance comparison of our models. In this experiment, we train and test the models in different domains. }
\end{table}

Although 30k reviews with rating tags are generally easy to obtain on online platforms, there will still be some domains (like unpopular products) where it is difficult to obtain such a large amount of data. Therefore, it is necessary to use the document-level sentiment classification data in other domains to complete the training of the model and apply it directly to the target domain. Therefore, we test the performance of our model under the cross-domain setting on SemEval2014 datasets, which means we train our model on the yelp-large dataset then test on the Lap14 dataset and train our model on the electronics-large dataset the test on the Res14 dataset. It can be seen in the Table~\ref{table:Cross-Domain-Analysis-table} that the performance of our model under the cross-domain setting does not decline too much compared the results under the in-domain setting. And in the laptop domain, the accuracy and macro-F1 score under cross-domain setting even exceeds in-domain setting. This shows that our model has a good cross-domain ability.

\subsection{Results on ARTS}


\begin{table}[ht]
	\begin{center}
	\begin{tabular}{ccccc}
		\hline
		\multirow{2}{*}{Model} & \multicolumn{2}{c}{ARTS-Res} & \multicolumn{2}{c}{ARTS-Lap} \\ \cline{2-5} 
		& Acc.           & F1           & Acc.           & F1           \\ \hline
		BERT-sent              & 37.39         & 35.94        & 36.78         & 34.46        \\ \hline
		AF-DSC                 & 39.07         & 38.34        & 39.30         & 36.74        \\ \hline
	\end{tabular}
	\end{center}
	\caption{\label{table:Results-on-ARTS-table} The results of BERT-sent and AF-DSC on ARTS \cite{Xing:2020}. }
\end{table}

To show that our AF-DSC is different from directly using the document level sentiment as aspect sentiment, we test AF-DSC and BERT-sent on the ARTS dataset. From Table~\ref{table:Results-on-ARTS-table}, our AF-DSC model exceeds the BERT-sent model, which shows that although our model is trained with document-level sentiment classification data, our method can better distinguish the sentiment polarity of the aspects in the review
compared with the BERT-sent model. However, it is not difficult to see that the performance achieved by our model is still low, which also shows that our model still is affected by the overall sentiment of the review. In future work, we will seek more appropriate methods to enhance the ability of the model to distinguish aspect sentiment.

\subsection{Ablation Study}

\begin{table}[ht]
	\begin{center}
	\begin{tabular}{lllll}
		\hline
		\multicolumn{1}{l}{Model} & \multicolumn{2}{l}{Res14} & \multicolumn{2}{l}{Lap14} \\ \cline{2-5} 
		\multicolumn{1}{l}{}      & Acc.         & F1          & Acc.         & F1          \\ \hline
		AF-DSC               & \textbf{69.61}       & \textbf{61.69}       & \textbf{65.41}       & 57.81       \\ \hline
		-wsp                      & 68.99       & 60.81       & 65.19       & \textbf{58.17}       \\ \hline
		-mwp                      & 68.78       & 60.96       & 64.61       & 57.27       \\ \hline
		-POS mask              & 62.50       & 58.15       & 63.67       & 54.16       \\ \hline
	\end{tabular}
	\end{center}
	\caption{\label{table:Ablation-Study-table} Ablation test of our model. -wsp/-mwp means we remove the word sentiment prediction/mask sentiment prediction task respectively. And -POS mask means that we consider non-noun words when we perform the attention operation.}
\end{table}

We conduct an ablation study to investigate the influence of different training tasks in our model. We remove the Word Sentiment Prediction and Mask Word Prediction respectively. The experiment in Table~\ref{table:Ablation-Study-table} shows that both of them contribute to the performance of our model.

Under the setting of -POS mask, we have $ {\beta _i} = {\widehat \beta _i} $, so that the model can pay attention to the hidden state of all words in the attention layer. The performance of the model degrades. A reasonable explanation is that due to the lack of the POS mask, the model tends to directly focus on those sentiment words, and ignores the establishment of the relationship between the sentiment words and aspects.

\subsection{Comparison of Different Document Sentiment Representation Methods}

\begin{table}[ht]
	\begin{center}
	\begin{tabular}{lcccc}
    \hline
    \multicolumn{1}{c}{\multirow{2}{*}{Method}} & \multicolumn{2}{c}{Res14}       & \multicolumn{2}{c}{Lap14}       \\ \cline{2-5} 
    \multicolumn{1}{c}{}                        & Acc.           & F1             & ACC            & F1             \\ \hline
    POS+ATT                                 & \textbf{69.61} & \textbf{61.69} & \textbf{65.41} & \textbf{57.81} \\ \hline
    Pool\_rep                                   & 60.38          & 55.80          & 63.04          & 54.21          \\ \hline
    AVG                                         & 60.04          & 56.33          & 61.14          & 53.28          \\ \hline
    \end{tabular}
	\end{center}
	\caption{\label{table:Different-Representation-table} Effects of different document-level sentiment representation methods on the performance of the model. }
\end{table}

In this section, we explore the impact of different kinds of document-level sentiment representation methods on the zero-shot aspect-based sentiment classification performance of our model. As shown in Table~\ref{table:Different-Representation-table}, POS+ATT is the method we used in the AF-DSC model. In Pool\_rep, we take the hidden state corresponding to [CLS] as the document-level sentiment representation of review; And in AVG, we average all the hidden states as the document-level sentiment representation of review. Pool\_Rep and AVG are commonly used to form the document-level sentiment representation in previous work. We can find that the performance of these methods decreases significantly, which shows that these methods are difficult to make the model actively establish the relationship between the sentiment polarity words and aspect. It is hard to transfer the sentiment information in the sentiment polarity words to the representation of ASC-related units.

\subsection{Case Study}

\begin{table*}
	\begin{center}
	\begin{tabular}{cp{6cm}ccc}
		\hline
		Dataset  & Review                                                                                        & Truth & BERT-sent & AF-DSC \\ \hline
		Restaurants &
		So we sat at the \underline{bar}, the \underline{bartender} did n't seem like he wanted to be there . &
		\begin{tabular}[c]{@{}c@{}}Neu\\ Neg\end{tabular} &
		\begin{tabular}[c]{@{}c@{}}Neg\\ Neg\end{tabular} &
		\begin{tabular}[c]{@{}c@{}}Neu\\ Neg\end{tabular} \\ \hline
		Laptops  & The Mountain Lion OS is not hard to   figure out if you are familiar with \underline{Microsoft Windows}. & Neu   & Neg       & Neu    \\ \hline
		ARTS-Res & The food is top   notch, but the \underline{service} is heedless, even if the atmosphere is great.        & Neg   & Neu       & Neg    \\ \hline
	\end{tabular}
	\end{center}
	\caption{\label{table:Case-Study-table} We select some cases from different datasets to illustrate the effectiveness of our model. }
\end{table*}

In this section, to intuitively explain the effectiveness of our model, we have selected some cases from the datasets. Those cases are listed on Table~\ref{table:Case-Study-table}.

In the first case, there are two aspects ``bar'' and ``bartender''. Because the BERT-sent model can not distinguish the aspects in the review, both of them are predicted to be negative. Our model seems can distinguish the sentiment polarity of the aspect, so it correctly predicts the different sentiment polarities of the two aspects.

In the second case, BERT-sent may be affected by ``not'' and ``hard'', and mistakenly judges the aspect sentiment as negative, while our AF-DSC model seems not to be affected by the two words and correctly classifies the results. This use case also reflects that the model is difficult in dealing with double negation.

The third case is selected from the ARTS dataset. It can be seen that the sentiment polarity of the target aspect is opposite to that of the surrounding aspects, which is very challenging for the model, especially for the model trained with document-level sentiment classification data. However, our AF-DSC model correctly classifies it, which shows that our model can capture the relationship between the aspect and sentiment polarity word. The neutral prediction result of the BERT-sent model is likely due to the positive and negative aspects simultaneously appealing in the review. The combination of them makes the document-level sentiment more inclined to be three stars, so it is neutral when converted to aspect sentiment.

\section{Conclusion}

In this paper, we explicitly clarify the semantic combination relation between DSC and ASC that the document-level sentiment representation is composed of ASC-related unit representations which can determine the sentiment of aspect and use this relationship to integrate the document-level sentiment label (rating) into the zero-shot aspect-based sentiment classification task. And we also design several training tasks to enhance the performance of our model. Compared to previous work, our model achieves better results. Experiments show that our model has a good cross-domain ability, and our model is different from directly using the document-level sentiment as the aspect sentiment and has certain characteristics required by aspect-based sentiment classification which is identifying the different sentiment polarities of different aspects in a review.


\bibliographystyle{acl_natbib}
\bibliography{anthology,custom}

\begin{thebibliography}{19}
\expandafter\ifx\csname natexlab\endcsname\relax\def\natexlab#1{#1}\fi

\bibitem[{Chen and Qian(2019)}]{Chen:2019}
Zhuang Chen and Tieyun Qian. 2019.
\newblock \href {https://doi.org/10.18653/v1/P19-1052} {Transfer capsule
  network for aspect level sentiment classification}.
\newblock In \emph{Proceedings of the 57th Annual Meeting of the Association
  for Computational Linguistics}, pages 547--556, Florence, Italy. Association
  for Computational Linguistics.

\bibitem[{He et~al.(2018)He, Lee, Ng, and Dahlmeier}]{He:2018}
Ruidan He, Wee~Sun Lee, Hwee~Tou Ng, and Daniel Dahlmeier. 2018.
\newblock \href {https://doi.org/10.18653/v1/P18-2092} {Exploiting document
  knowledge for aspect-level sentiment classification}.
\newblock In \emph{Proceedings of the 56th Annual Meeting of the Association
  for Computational Linguistics (Volume 2: Short Papers)}, pages 579--585,
  Melbourne, Australia. Association for Computational Linguistics.

\bibitem[{He et~al.(2019)He, Lee, Ng, and Dahlmeier}]{He:2019}
Ruidan He, Wee~Sun Lee, Hwee~Tou Ng, and Daniel Dahlmeier. 2019.
\newblock \href {https://doi.org/10.18653/v1/P19-1048} {An interactive
  multi-task learning network for end-to-end aspect-based sentiment analysis}.
\newblock In \emph{Proceedings of the 57th Annual Meeting of the Association
  for Computational Linguistics}, pages 504--515, Florence, Italy. Association
  for Computational Linguistics.

\bibitem[{Honnibal and Montani(2017)}]{Honnibal:2017}
Matthew Honnibal and Ines Montani. 2017.
\newblock {spaCy 2}: Natural language understanding with {B}loom embeddings,
  convolutional neural networks and incremental parsing.
\newblock To appear.

\bibitem[{Hu and Liu(2004)}]{Hu:2004}
Minqing Hu and Bing Liu. 2004.
\newblock Mining and summarizing customer reviews.
\newblock In \emph{Proceedings of the tenth ACM SIGKDD international conference
  on Knowledge discovery and data mining}, pages 168--177.

\bibitem[{Ke et~al.(2020)Ke, Ji, Liu, Zhu, and Huang}]{Ke:2020}
Pei Ke, Haozhe Ji, Siyang Liu, Xiaoyan Zhu, and Minlie Huang. 2020.
\newblock \href {https://doi.org/10.18653/v1/2020.emnlp-main.567}
  {{S}enti{LARE}: Sentiment-aware language representation learning with
  linguistic knowledge}.
\newblock In \emph{Proceedings of the 2020 Conference on Empirical Methods in
  Natural Language Processing (EMNLP)}, pages 6975--6988, Online. Association
  for Computational Linguistics.

\bibitem[{Li et~al.(2019)Li, Wei, Zhang, Zhang, and Li}]{Li:2019}
Zheng Li, Ying Wei, Yu~Zhang, Xiang Zhang, and Xin Li. 2019.
\newblock Exploiting coarse-to-fine task transfer for aspect-level sentiment
  classification.
\newblock In \emph{Proceedings of the AAAI Conference on Artificial
  Intelligence}, volume~33, pages 4253--4260.

\bibitem[{Liu(2012)}]{Liu:2012}
Bing Liu. 2012.
\newblock Sentiment analysis and opinion mining.
\newblock \emph{Synthesis lectures on human language technologies},
  5(1):1--167.

\bibitem[{Pang et~al.(2008)Pang, Lee et~al.}]{Pang:2008}
Bo~Pang, Lillian Lee, et~al. 2008.
\newblock Opinion mining and sentiment analysis.
\newblock \emph{Foundations and Trends{\textregistered} in information
  retrieval}, 2(1--2):1--135.

\bibitem[{Pontiki et~al.(2014)Pontiki, Galanis, Pavlopoulos, Papageorgiou,
  Androutsopoulos, and Manandhar}]{Pontiki:2014}
Maria Pontiki, Dimitris Galanis, John Pavlopoulos, Harris Papageorgiou, Ion
  Androutsopoulos, and Suresh Manandhar. 2014.
\newblock \href {https://doi.org/10.3115/v1/S14-2004} {{S}em{E}val-2014 task 4:
  Aspect based sentiment analysis}.
\newblock In \emph{Proceedings of the 8th International Workshop on Semantic
  Evaluation ({S}em{E}val 2014)}, pages 27--35, Dublin, Ireland. Association
  for Computational Linguistics.

\bibitem[{Sainz et~al.(2021)Sainz, Lopez~de Lacalle, Labaka, Barrena, and
  Agirre}]{Sainz:2021}
Oscar Sainz, Oier Lopez~de Lacalle, Gorka Labaka, Ander Barrena, and Eneko
  Agirre. 2021.
\newblock \href {https://doi.org/10.18653/v1/2021.emnlp-main.92} {Label
  verbalization and entailment for effective zero and few-shot relation
  extraction}.
\newblock In \emph{Proceedings of the 2021 Conference on Empirical Methods in
  Natural Language Processing}, pages 1199--1212, Online and Punta Cana,
  Dominican Republic. Association for Computational Linguistics.

\bibitem[{Seoh et~al.(2021)Seoh, Birle, Tak, Chang, Pinette, and
  Hough}]{Seoh:2021}
Ronald Seoh, Ian Birle, Mrinal Tak, Haw-Shiuan Chang, Brian Pinette, and Alfred
  Hough. 2021.
\newblock \href {https://doi.org/10.18653/v1/2021.emnlp-main.509} {Open aspect
  target sentiment classification with natural language prompts}.
\newblock In \emph{Proceedings of the 2021 Conference on Empirical Methods in
  Natural Language Processing}, pages 6311--6322, Online and Punta Cana,
  Dominican Republic. Association for Computational Linguistics.

\bibitem[{Shu et~al.(2022)Shu, Chen, Liu, and Xu}]{Shu:2022}
Lei Shu, Jiahua Chen, Bing Liu, and Hu~Xu. 2022.
\newblock Zero-shot aspect-based sentiment analysis.
\newblock \emph{arXiv preprint arXiv:2202.01924}.

\bibitem[{Sun et~al.(2019)Sun, Huang, and Qiu}]{Sun:2019}
Chi Sun, Luyao Huang, and Xipeng Qiu. 2019.
\newblock \href {https://doi.org/10.18653/v1/N19-1035} {Utilizing {BERT} for
  aspect-based sentiment analysis via constructing auxiliary sentence}.
\newblock In \emph{Proceedings of the 2019 Conference of the North {A}merican
  Chapter of the Association for Computational Linguistics: Human Language
  Technologies, Volume 1 (Long and Short Papers)}, pages 380--385, Minneapolis,
  Minnesota. Association for Computational Linguistics.

\bibitem[{Xing et~al.(2020)Xing, Jin, Jin, Wang, Zhang, and Huang}]{Xing:2020}
Xiaoyu Xing, Zhijing Jin, Di~Jin, Bingning Wang, Qi~Zhang, and Xuanjing Huang.
  2020.
\newblock \href {https://doi.org/10.18653/v1/2020.emnlp-main.292} {Tasty
  burgers, soggy fries: Probing aspect robustness in aspect-based sentiment
  analysis}.
\newblock In \emph{Proceedings of the 2020 Conference on Empirical Methods in
  Natural Language Processing (EMNLP)}, pages 3594--3605, Online. Association
  for Computational Linguistics.

\bibitem[{Xu et~al.(2019)Xu, Liu, Shu, and Yu}]{Xu:2019}
Hu~Xu, Bing Liu, Lei Shu, and Philip Yu. 2019.
\newblock \href {https://doi.org/10.18653/v1/N19-1242} {{BERT} post-training
  for review reading comprehension and aspect-based sentiment analysis}.
\newblock In \emph{Proceedings of the 2019 Conference of the North {A}merican
  Chapter of the Association for Computational Linguistics: Human Language
  Technologies, Volume 1 (Long and Short Papers)}, pages 2324--2335,
  Minneapolis, Minnesota. Association for Computational Linguistics.

\bibitem[{Yin et~al.(2019)Yin, Hay, and Roth}]{Yin:2019}
Wenpeng Yin, Jamaal Hay, and Dan Roth. 2019.
\newblock \href {https://doi.org/10.18653/v1/D19-1404} {Benchmarking zero-shot
  text classification: Datasets, evaluation and entailment approach}.
\newblock In \emph{Proceedings of the 2019 Conference on Empirical Methods in
  Natural Language Processing and the 9th International Joint Conference on
  Natural Language Processing (EMNLP-IJCNLP)}, pages 3914--3923, Hong Kong,
  China. Association for Computational Linguistics.

\bibitem[{Yin et~al.(2020)Yin, Rajani, Radev, Socher, and Xiong}]{Yin:2020}
Wenpeng Yin, Nazneen~Fatema Rajani, Dragomir Radev, Richard Socher, and Caiming
  Xiong. 2020.
\newblock \href {https://doi.org/10.18653/v1/2020.emnlp-main.660} {Universal
  natural language processing with limited annotations: Try few-shot textual
  entailment as a start}.
\newblock In \emph{Proceedings of the 2020 Conference on Empirical Methods in
  Natural Language Processing (EMNLP)}, pages 8229--8239, Online. Association
  for Computational Linguistics.

\bibitem[{Zhou et~al.(2020)Zhou, Tian, Wang, Wu, Xiao, and He}]{Zhou:2020}
Jie Zhou, Junfeng Tian, Rui Wang, Yuanbin Wu, Wenming Xiao, and Liang He. 2020.
\newblock \href {https://doi.org/10.18653/v1/2020.coling-main.49} {{S}enti{X}:
  A sentiment-aware pre-trained model for cross-domain sentiment analysis}.
\newblock In \emph{Proceedings of the 28th International Conference on
  Computational Linguistics}, pages 568--579, Barcelona, Spain (Online).
  International Committee on Computational Linguistics.

\end{thebibliography}

\appendix



\end{document}